\documentclass[letterpaper, 10 pt, conference]{ieeeconf}  

\IEEEoverridecommandlockouts                              

\usepackage[T1]{fontenc}
\usepackage{blindtext}
\usepackage{graphicx}
\usepackage{xcolor}

\usepackage{amsmath}
\usepackage{amssymb}

\usepackage{graphicx}
\usepackage{placeins}

\usepackage{subfig}
\usepackage{gensymb}
\usepackage{textcomp}
\usepackage{hyperref}
\usepackage{amsmath}

\DeclareMathOperator{\acos}{acos}

\usepackage{multirow}
\usepackage{balance}
\usepackage{bm}

\usepackage{amsmath}

\DeclareMathOperator*{\argmin}{arg\,min}
\usepackage{mathtools}

\usepackage{lipsum}
\usepackage[colorinlistoftodos]{todonotes} 

\usepackage{tikz}
\usepackage{textcomp}
\usepackage{hyperref}
\usepackage{lipsum}

\newcommand\copyrighttext{%
  \footnotesize \textcopyright 2020 IEEE. Personal use of this material is permitted.
  Permission from IEEE must be obtained for all other uses, in any current or future
  media, including reprinting/republishing this material for advertising or promotional
  purposes, creating new collective works, for resale or redistribution to servers or
  lists, or reuse of any copyrighted component of this work in other works.
  DOI: \href{https://ieeexplore.ieee.org/document/9161262}{10.1109/LRA.2020.3014639}}
\newcommand\copyrightnotice{%
\begin{tikzpicture}[remember picture,overlay]
\node[anchor=south,yshift=10pt] at (current page.south) {\fbox{\parbox{\dimexpr\textwidth-\fboxsep-\fboxrule\relax}{\copyrighttext}}};
\end{tikzpicture}%
}

\title{\LARGE \bf
Spatiotemporal Calibration of Camera and 3D Laser Scanner
}

\author{Micha\l{} R. Nowicki$^{1}$
\thanks{This work was supported by the National Center for Research and Development under the grant POIR.04.01.02-00-0081/17}
\thanks{The helmet-based sensory system used in this paper was developed by J. Wietrzykowski within the H2020 730994 TERRINet grant SMILE}
\thanks{$^{1}$ The author is with the Institute of Robotics and Machine Intelligence,
        Poznan University of Technology, Poznan, Poland
        {\tt\small michal.nowicki@put.poznan.pl}}%
}

\begin{document}

\maketitle

\copyrightnotice

\thispagestyle{empty}
\pagestyle{empty}

\begin{abstract}
The multi-sensory setups consisting of the laser scanners and cameras are popular as the measurements complement each other and provide necessary robustness for applications. 
Under dynamic conditions or when in motion, a direct transformation (spatial calibration) and time offset between sensors (temporal calibration) is needed to determine the correspondence between measurements.
We propose an open-source spatiotemporal calibration framework for a camera and a 3D laser scanner. 
Our solution is based on commonly available chessboard markers requiring one-minute calibration before the operation that offers accurate and repeatable results.
The framework is based on batch optimization of point-to-plane constraints with a time offset calibration possible by a novel continuous representation of the plane equations based on a minimal representation in the Lie algebra and the use of B-splines.
The framework's properties are evaluated in simulation while correctness is verified with two distinct sensory setups with Velodyne VLP-16 and SICK MRS6124 3D laser scanners.

\end{abstract}

\section{Introduction}

Nowadays, the sensory setups of mobile robots consist of an increasing number of cameras, LiDARs, radars, and AHRS units to provide complementary information used by autonomous robots. 
To use different sensors for the state estimation it is necessary to represent the measurement from one sensor in the coordinate system of the other sensor.
Under static conditions, only the spatial transformation is needed.
But when sensors are in motion (or when the environment is changing), the knowledge on the temporal relationship between sensors is necessary.

The spatial calibration can be estimated with already existing frameworks or assumed based on CAD drawings.
The temporal calibration is usually performed with the additional hardware that triggers cameras and timestamps acquired 3D laser scans.
This additional hardware is often not available out-of-the-box and some cheaper versions of sensors do not offer such possibility at all.

\begin{figure}[htbp!]
    \centering
    \includegraphics[width=\columnwidth]{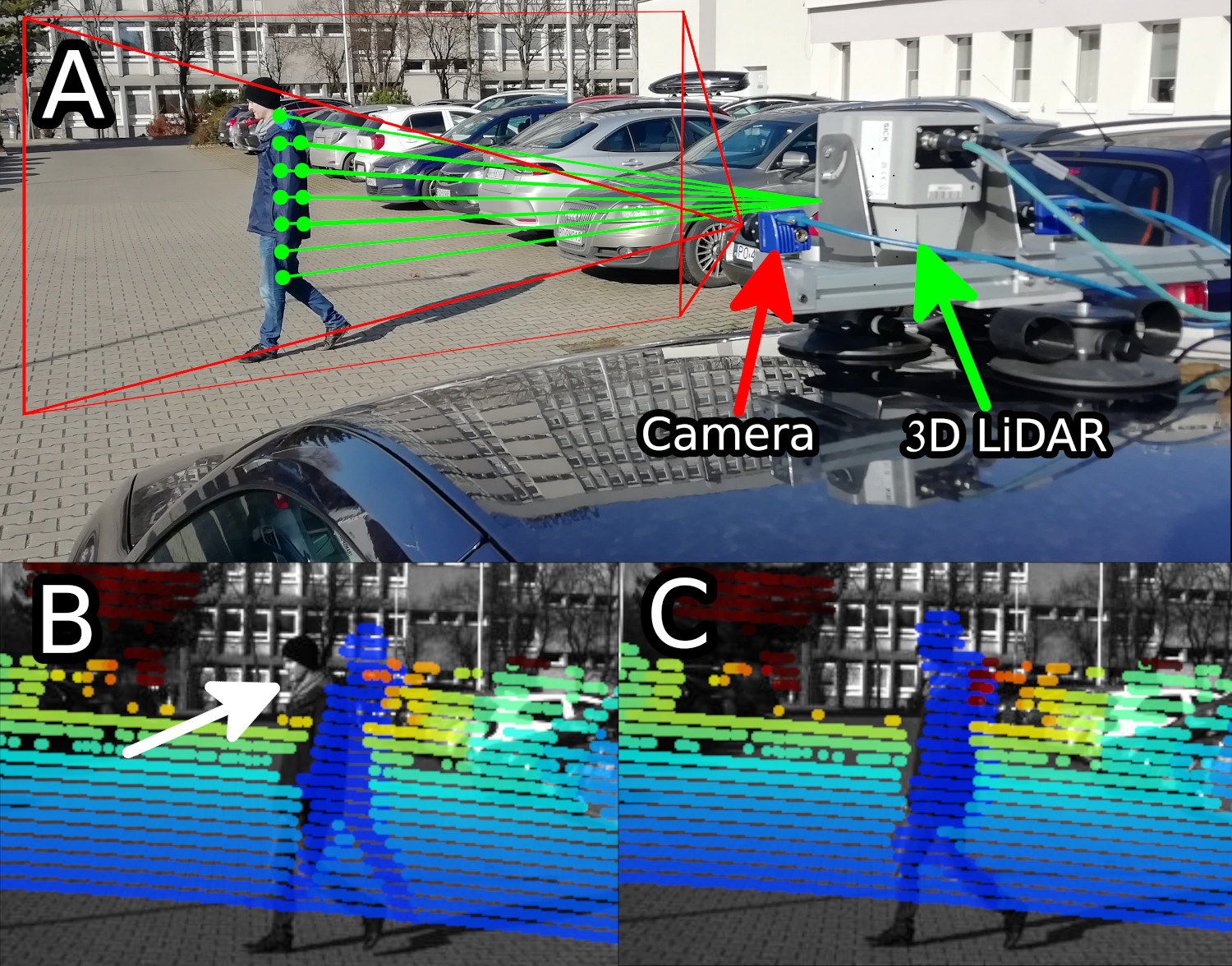}
     \caption{Common observation from a camera-3D LiDAR sensory setup mounted on a car (A) and LiDAR points projected onto the image in a dynamic scenario when spatial (B) or spatiotemporal (C) calibration was performed with the proposed framework. Notice how lack of temporal calibration leads to incorrect sensor correspondence (white arrow)}
    \label{fig:catchy_image}
\end{figure}

In our work, we target the problem of joint spatiotemporal calibration of a sensory system consisting of a camera and a 3D laser scanner (LiDAR), as presented in Fig.~\ref{fig:catchy_image}A.
The spatial calibration provides a possibility to determine the correspondence between measurements from sensors under static conditions but fails in a dynamic environment (Fig.~\ref{fig:catchy_image}B) while spatiotemporal calibration provides expected results (Fig.~\ref{fig:catchy_image}C).
Spatiotemporal calibration is a necessity for currently popular tightly integrated Visual Odometry (VO) or Simultaneous Localization and Mapping (SLAM) solutions that jointly process data from a camera and 3D LiDAR~\cite{tightvo}.

The proposed calibration is based on the motion of a chessboard pattern and a sequence of successive observations from the camera and 3D LiDAR.
The system performs batch optimization of constraints to estimate 6 DoF (degree-of-freedom) transformation between sensors and the corresponding time offset.
The optimization is based on independent LiDAR's point to the camera's plane constraints that are extended by the continuous-time plane representation to allow time offset optimization.

The contribution of our work can be summarized as:
\begin{itemize}
    \item spatiotemporal camera - 3D laser scanner calibration that makes it possible to estimate 6 DoF transformation and the time offset between sensors using common chessboard marker,
    \item novel continuous-time minimal plane representation using the Lie algebra with B-splines,
    \item calibration solution that converges to expected values even with poor spatial and temporal initial guesses,
    \item publicly available, ROS-compatible calibration  framework\footnote{\url{https://github.com/LRMPUT/CameraLidarCalibrator}}.
\end{itemize}

\section{Related work}

The need for multisensor calibration can be dated back to camera-2D laser sensory rigs that were used to design multisensor SLAM systems~\cite{tardos}.
The first popular and publicly available framework for spatial calibration between a camera and a 2D laser scanner was presented by Zhang and Pless in~\cite{zhang04}.
The calibration was a result of an optimization performed over several observations from both sensors observing a camera calibration pattern in discrete calibration poses. 
The 2D laser points lying on the camera calibration plane were used to form and solve the non-linear optimization problem with point-to-plane constraints.

The first spatial calibration of a camera and 3D LiDAR was presented by Scaramuzza in~\cite{sca07}.  This solution was based on manual markings between corresponding points between the camera's image and LiDAR's scan converted to range images and was suited to rotating 2D laser scanners under static conditions that could generate high-resolution range images. 

The spatial calibration of a camera and 3D LiDAR was later tackled with the use of specially modified calibration patterns with four circular holes~\cite{velas}, a box with 3 perpendicular sides (trihedron)~\cite{trihedron}, or just regular boxes~\cite{boxes}. 
Despite different approaches to detection, constraint extraction, and optimization formulation, none of these approaches provides temporal calibration.

More recently, another branch of solutions emerged with an idea to perform calibration based on a comparison between the independent trajectories from both (or multiple) analyzed sensors.
These solutions can also be based on tracking of some object (marker), like a ball of a certain color in the calibration of multiple RGB-D Kinect v2 sensors~\cite{fornaser}, or marker-less, like proposed by~\cite{ishi18} that focuses on working with short sequences or~\cite{taylorTRO} that considers sensor's accuracy to provide the uncertainty of the obtained calibration.
In~\cite{persic}, the authors focus on the trajectory representation proposing Gaussian processes as a better alternative for trajectory alignment.
The marker-less solutions do not rely on the placement of an artificial marker in the environment but the experimental environment is expected to be well-suited for VO/SLAM systems.

The spatiotemporal calibration with those approaches is usually performed by decoupling the problem of spatial and temporal calibration.
The temporal calibration is often found with some form of cross-correlation on the time-evolution of a signal that is independent of the spatial calibration (like linear or rotational speeds)~\cite{taylorTRO,persic}. 
Some researchers believe that these estimations should not be decoupled~\cite{kalibr1} and the problem should be solved simultaneously.
More recently, Park et al.~\cite{park20} proposed a solution that utilizes linear interpolation on the $SE(3)$ Lie group to jointly solve the spatiotemporal problem for marker-less calibration.

We propose a spatiotemporal calibration that is marker-based and relies on continuous relative motion between the sensory setup and the marker. 
The continuous motion provides us with a significantly bigger number of measurements than the static calibration while a novel continuous-time representation of plane equations in the Lie algebra with B-splines makes it possible to estimate time offsets between sensors.
Compared to marker-less solutions, our marker-based solution is independent of the accuracy of the VO/SLAM or the type of the environment, while also providing both spatial and temporal calibration.
Due to the continuous nature of data acquisition, the proposed calibration is similar to the~\textit{kalibr}~\cite{kalibr2} that offers spatiotemporal calibration for multi-camera and multi-AHRS sensory setups, where time-varying states are represented by B-spline functions, as introduced in~\cite{prekalibr}.
Unfortunately, \textit{kalibr} only provides an experimental camera-LRF (2D LiDAR) calibration that is no longer maintained in the public repository.
Therefore, there is a need for our contribution that can be combined with \textit{kalibr} to calibrate more complex sensory setups.


\section{Proposed solution}

\begin{figure*}[h!]
    \centering
    \includegraphics[width=\textwidth]{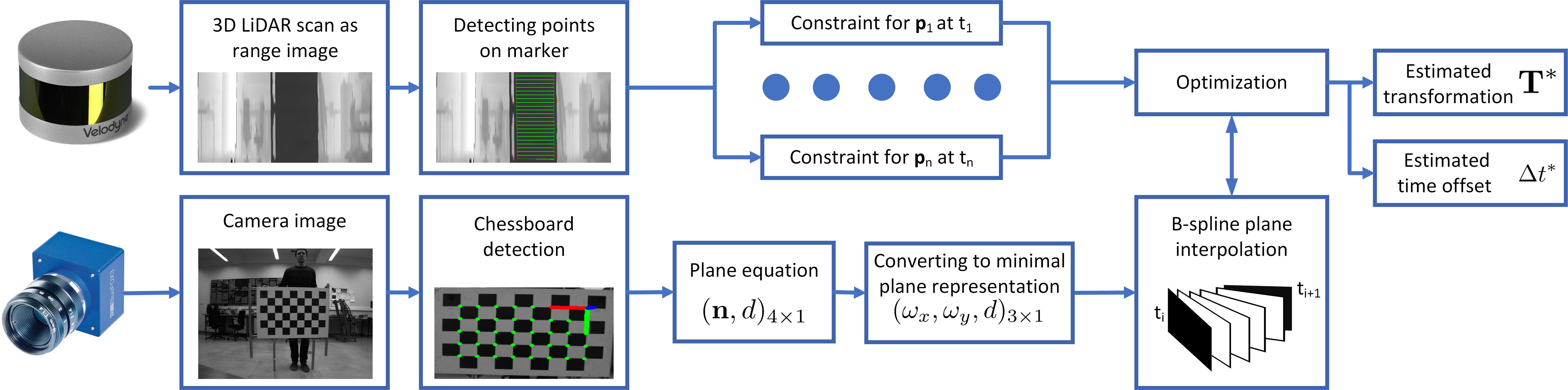}
    \caption{The overall processing steps of the proposed spatiotemporal camera-3D LiDAR calibration}
    \label{fig:proc_steps}
\end{figure*}

The proposed framework is dedicated to sensory setups with global-shutter cameras and 3D LiDARs that are rigidly attached. 
We assume that the camera is internally calibrated and these parameters are fixed during the calibration.
Besides, we assume that the time offset between sensors is unknown, small, and is approximately constant.
This assumption is true in most cases, especially when sensors are connected to the same computer.

The calibration procedure consists of motion of a calibration pattern that should be jointly and continuously observed by the camera and by the 3D LiDAR. 
Depending on the preference of the user, either the sensory setup or the calibration pattern can be moved.
After recording, we process the data as presented in Fig.~\ref{fig:proc_steps}.
The calibration pattern is detected on the camera images using typical chessboard detection algorithm from OpenCV.
The LiDAR points belonging to the plane of the calibration pattern are detected using range images either by manual marking of the corners or with a semi-automatic tracking from the previous detection.
Without temporal calibration, the spatial calibration can be determined by the point-to-plane optimization based on a discrete number of calibration poses:
\begin{equation}
{\mathbf T}^* = \argmin_{\mathbf T} \sum_i \sum_j {\bm \pi}(t_i)^{\top} {\mathbf T}  {\mathbf p}_i,
\end{equation}
where ${\mathbf T}^*$ is the sought transformation from the coordinate system of the LiDAR to the coordinate system of the camera, ${\mathbf p}_i$ denotes the homogeneous coordinates of the $i$-th 3D laser point on the calibration pattern, and ${\bm \pi}_(t_i)$ is the plane equation of the chessboard pattern estimated by the camera at timestamp $(t_i + \Delta t)$.
The plane ${\bm \pi}_{(t_i)}$ can be determined by finding the plane equation for the detection on the camera image with the closest timestamp.
The frequency of captured data from both systems can be different in the presented formulation.

In the proposed framework, we are also concerned with temporal calibration, which raises several issues as data from both sensors are not recorded in the same moments (timestamps) and 3D LiDARs are constantly rotating, which results in motion distortions similar to those in rolling-shutter cameras.
These distortions can be compensated to create a point cloud at selected timestamp within SLAM solutions, like for 3D LiDARs in~\cite{loam} or cameras in~\cite{loveskew}, but that inevitably introduces additional errors.
We opt to treat each point with the corresponding timestamp independently, similarly to the first odometry steps in~\cite{loam} or in~\cite{zebedee}.
In this more complicated case, the spatiotemporal calibration is determined by the following optimization:
\begin{equation}
\label{eq:our}
    {\mathbf T}^*, \Delta t^* = \argmin_{{\mathbf T}, \Delta t} \sum_i {\bm \pi}(t_i + \Delta t)^{\top} {\mathbf T} {\mathbf p}_i,
\end{equation}
where $\Delta t^*$ is the estimated time difference between time\-stamps of the camera and the 3D LiDAR, ${\bm \pi}(t_i + \Delta t)$ is the plane equation of the chessboard pattern estimated by the camera at timestamp $(t_i + \Delta t)$, and $t_i$ is the acquisition timestamp of the $i$-th laser point ${\mathbf p}_i$ lying on the plane of the calibration pattern.
Formulation of the problem as in (\ref{eq:our}) requires individual timestamps for each 3D laser point and the knowledge of the plane equation at any arbitrary timestamp. 
The optimization is performed using the Levenberg-Marquardt algorithm with the Huber robust cost available in the \textit{g2o} library~\cite{g2o}.


\subsection{Time offsets for points from 3D laser scanner}

Most of the available 3D laser scanners operate as continuously rotating devices that measure the distance based on the time-of-flight principle. 
As each device is constructed differently, the direct timestamps for each laser point have to be determined.
The best calibration results should be expected if sensor drivers provide real timestamps for all considered points. 
However, in some case it is not possible and the real laser point timestamps have to be estimated using the available information about the method of acquiring individual measurements in the particular LiDAR type.
We present laser point timestamps estimation for Velodyne VLP-16 and SICK MRS6124 that have very different patterns of laser points acquisition.

\begin{figure}[htbp!]
    \centering
    \includegraphics[width=\columnwidth]{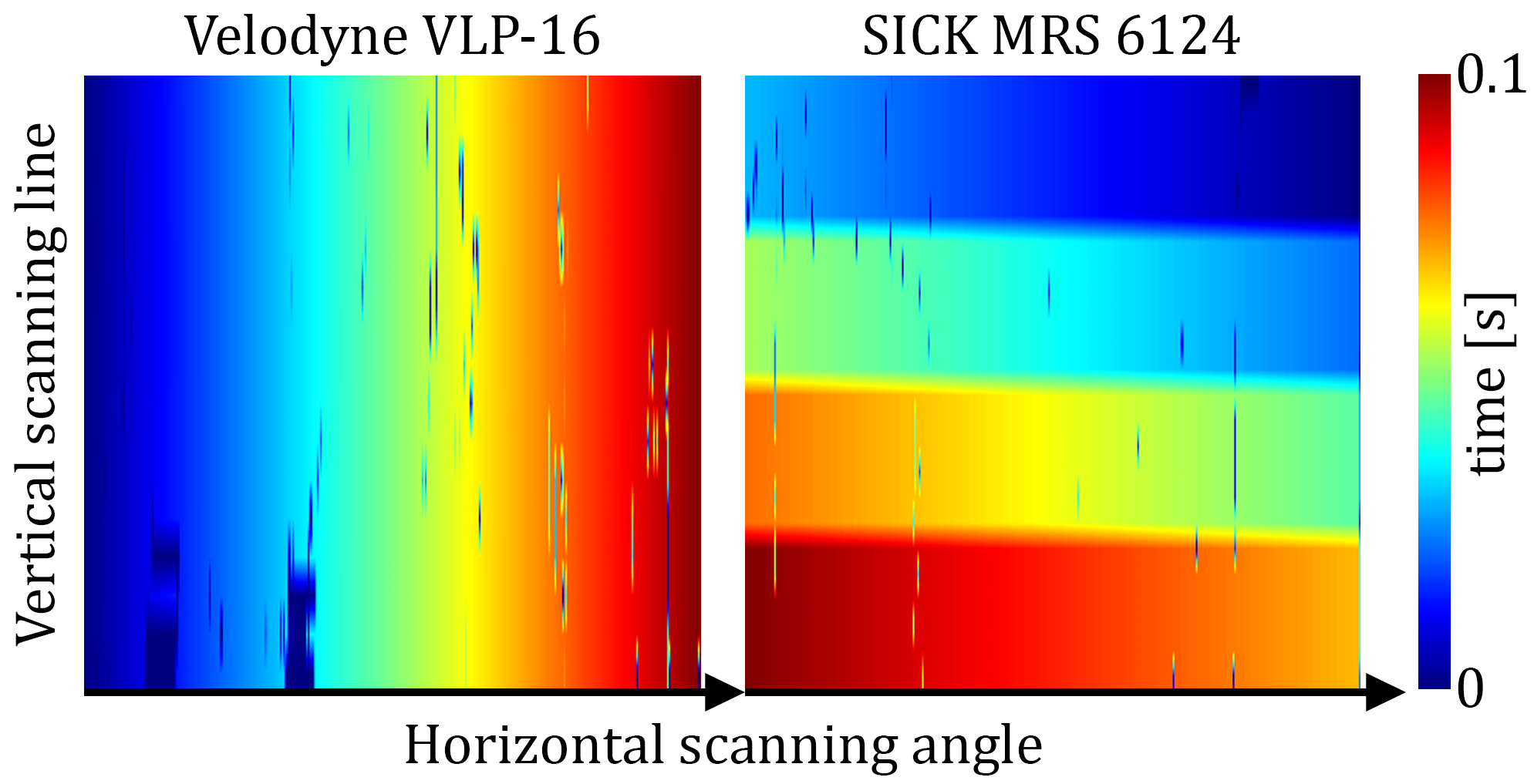}
    \caption{Time offsets for laser points belonging to Velodyne VLP-16 and SICK MRS6124 depending on the time of acquisition due to beam rotation. In real scans, some distance measurements are invalid and thus some missing time offsets can be observed visible as artifacts on the presented time offset images}
    \label{fig:laser_scanner}
\end{figure}

The Velodyne VLP-16 has 16 scanning layers, $360\degree$ field of view, and in the typical configuration rotates at $f$ = 10 Hz in the clockwise direction. 
For each retrieved point cloud, we determine the starting (lowest) angle $\phi_s$ and the ending (largest) angle $\phi_e$ as the angles on the horizontal scanning plane.
The time offset $t_i$ for the point ${\mathbf p}_i$ is computed as:
\begin{equation}
    t_i = t_{\rm cloud} + \frac{\phi_i - \phi_s}{f (\phi_e - \phi_s)},
\end{equation}
where $t_{\rm cloud}$ stands for the timestamp of the pointcloud and $\phi_i$ is the horizontal scanning angle for the analyzed point.

The SICK MRS6124 has 24 scanning layers but these layers are formed from 4 groups with 6 independent scanning lines.
The sensor has a $120\degree$ field of view and rotates in the counter-clockwise direction.
This scanner also provides data with $f$ = 10 Hz, but rotates 4 times faster as the 6 independent lines perform full scanning for each group one after the other.
In this case, the time offset $t_i$ for point $p_i$ is computed as:
\begin{equation}
    t_i = t_{\rm cloud} + \frac{1}{f} \left(\frac{g}{4} + \frac{\phi_i - \phi_s}{2\pi}\right), 
\end{equation}
where $g$ is the number of the scanning group (0, 1, 2 or 3) and $\phi_s = \frac{\pi}{3}$.

The time offsets for points in the same scan for both sensors are visualized in Fig.~\ref{fig:laser_scanner}.

\subsection{Continuous plane representation}

The chessboard pattern is detected in each incoming camera image. 
With the knowledge of the camera parameters and real chessboard size, the transformation between the camera coordinate system and coordinate system of the chessboard pattern is determined.
Then, the chessboard calibration plane equation is computed and represented by the 4-dimensional vector $(\mathbf{n}, d)$,
where $\mathbf{n}$ is the normalized three-dimensional plane normal, and $d$ is the distance to the origin of the coordinate system.
From all images, we get a set of plane equations observed at discrete timestamps of images.

To determine the plane equations at the timestamps between timestamps of images, we need to interpolate these plane equations. 
The 4-dimensional representation is not minimal and thus the interpolation might have resulted in plane equations that require further re-normalization.
To avoid that issue, we propose a minimal $3$-dimensional representation utilizing properties of the $SO(3)$ group (following the ideas in~\cite{bartoli}) and its corresponding Lie algebra elements.
The exponential map ($\exp$) transforms elements from Lie algebra tangential space ($\mathfrak{so}(3)$) to $SO(3)$ and its inverse logarithmic map $\log$ in order to perform inverse transformation (from $SO(3)$ to $\mathfrak{so}(3)$).
The idea is that only two parameters are needed to represent a normalized vector (or a point on unit hemisphere).
Therefore the presented idea is almost identical to spherical interpolation as presented in~\cite{spherical}.

The overparametrization of a default plane representation stems from the fact that the plane
normal $\mathbf{n}$ is normalized, and thus can be represented with two components of the Lie algebra
of the $SO(3)$ group $(({\mathbf n}, d)_{4\times1} \rightarrow (\omega_x, \omega_y, d)_{3\times1} )$:
\begin{align}
\theta &= \acos({\mathbf n}(2)), \\
\omega_x &= - {\mathbf n}(1) * \frac{\theta}{\sin(\theta)}, \mbox{~~~}
\omega_y = {\mathbf n}(0) * \frac{\theta}{\sin(\theta)}
\end{align}
Special care has to be taken when $\theta \rightarrow 0$ (we perform series expansion of
$\frac{\theta}{\sin(\theta)}$) and when $\theta \rightarrow \pi$ (we avoid the issue by
assuming equivalent plane representation ${\mathbf n}\prime =-{\mathbf n}$ and $d\prime=-d$).
If necessary, the original, 4-dimensional representation can be retrieved with
$((\omega_x, \omega_y, d)_{3\times1} \rightarrow ({\mathbf n}, d)_{4\times1} )$: 
\begin{align}
\mathbf{n} &= \exp([\omega_x, \omega_y, 0]) \begin{bmatrix} 0 \\ 0 \\ 1 \end{bmatrix}, 
\end{align}
where $\exp(\cdot)$ computes the Lie group representation based on the Lie algebra element.
We encourage readers to analyze the provided open-source code when in doubt.

\subsection{Plane equation interpolation}

The minimal plane representation makes it possible to determine plane equations at required timestamps by converting to the minimal representation, performing interpolation, and then returning to the usual 4-dimensional plane equation. 
For this task, we tried linear interpolation of parameters, but the lack of continuous derivatives is an important issue that results in optimization getting stuck in local minima.
Therefore, we propose to perform interpolation with cubic B-splines for continuous plane representations that guarantee continuous first and second derivatives of the minimal plane representation. 

In our formulation of the B-spline we use the cumulative form as presented in~\cite{splinecum}:
\begin{equation}
\label{eq:spline1}
    \mathbf{s}(t) = \mathbf{s}_0 B_0(t) + \sum_{i=1}^n (\mathbf{s}_{i} - \mathbf{s}_{i-1}) B_i(t), 
\end{equation}
where $\mathbf{s}(t)$ is the interpolated value at time $t$, $\mathbf{s}_i$ is the value of the $i$-th control point that is a known value from a measurement, $B_i(t)$ is the $i$-th component of the cumulative basis function, and $n$ is the order of the B-spline.
The same formulation for the Lie algebra elements can be written as~\cite{spline}:
\begin{equation}
\label{eq:spline2}
    \mathbf{r}(t) = \log \{ \exp ( \mathbf{r}_0 B_0(t) ) \prod_{i=1}^n \exp ( \boldsymbol{\Omega}_{i} B_i(t)) \}, 
\end{equation}
where $\mathbf{r}(t)$ is the interpolated value of the Lie algebra element at time $t$, $\mathbf{r}_i$ is the value of the $i$-th control point in the Lie algebra (known value for time $t_i$), $\boldsymbol{\Omega}_{i} = \log (\exp(\mathbf{r}_{i-1})^T \exp(\mathbf{r}_{i}))$ is the equivalent of a difference for Lie algebra elements.
We use cubic B-spline ($n = 4$) and thus the cumulative basis function is equal to:
\begin{equation}
    \mathbf{B}(u) = \frac{1}{6} \begin{bmatrix}
6 & 0 & 0 & 0 \\
5 & 3 & -3 & 1 \\
1 & 3 & 3 & -2 \\
0 & 0 & 0 & 1
\end{bmatrix}
\begin{bmatrix}
1 \\
u \\
u^2 \\
u^3
\end{bmatrix},
\end{equation}
where $u$ is the normalized time (from $0$ to $1$) for the considered interval between $t_2$ and $t_3$.
In practice, we also check that the timestamps of the detected chessboards ($t_1, t_2, t_3, t_4$) are uniformly distributed in time and do not perform interpolation if this condition is not satisfied.

In the proposed solution, with minimal plane representation $(\omega_x, \omega_y, d)$, we follow (\ref{eq:spline1}) for the interpolation of element $d$ and (\ref{eq:spline2}) for Lie algebra components $(\omega_x, \omega_y)$.
Importantly, the analytical Jacobians of the B-spline can be derived for the presented interpolation using the well-known equations from~\cite{spline}.
Our system uses more efficient analytical Jacobians presented in~\cite{sommer19}.

\section{Analysis of the calibration system's properties}

\subsection{Simulation}
We evaluated our system in a series of Monte-Carlo type simulation experiments.
Each experiment starts with the creation of a motion that resembles a real movement of the calibration marker in front of a sensory setup and lasts for 50 seconds.
We start by randomly sampling 11 poses of the calibration marker in the typical working area of a calibration (cuboid of size $8\times2\times4$ meters) with an angle between plane normal and camera optical axis not exceeding $90\degree$.
These poses are considered to be observed at timestamps of 0, 5, 10, ..., 50 s of the calibration.
To achieve continuous motion of the calibration marker in the environment between these poses, we interpolate positions of the four corners of the calibration marker using B-splines on the $SE(3)$ group.
Based on the motion of the chessboard marker, we simulate camera chessboard detections with the desired framerate ($10$ Hz) and compute LiDAR points that would lie on that calibration pattern assuming random camera-LiDAR transformation with $t_x\in(-1,1), t_y\in(-0.5, 0.5), t_z\in(-0.25, 0.25)$ meters and angular rotation up to 45\textdegree~by random axis. 
In the end, we achieve a set of chessboard detections at selected timestamps and a set of LiDAR points with corresponding timestamps.
The initial guess of transformation is computed as a modification to the real transformation with $t_x\in(-0.1,0.1), t_y\in(-0.1, 0.1), t_z\in(-0.1, 0.1)$ meters and angular rotation up to $22.5$\textdegree~by random axis. 
The time offset initial guess is set to 0.

\subsection{Influence of LiDAR's accuracy}

\begin{figure}[htbp!]
    \centering
    \includegraphics[width=\columnwidth]{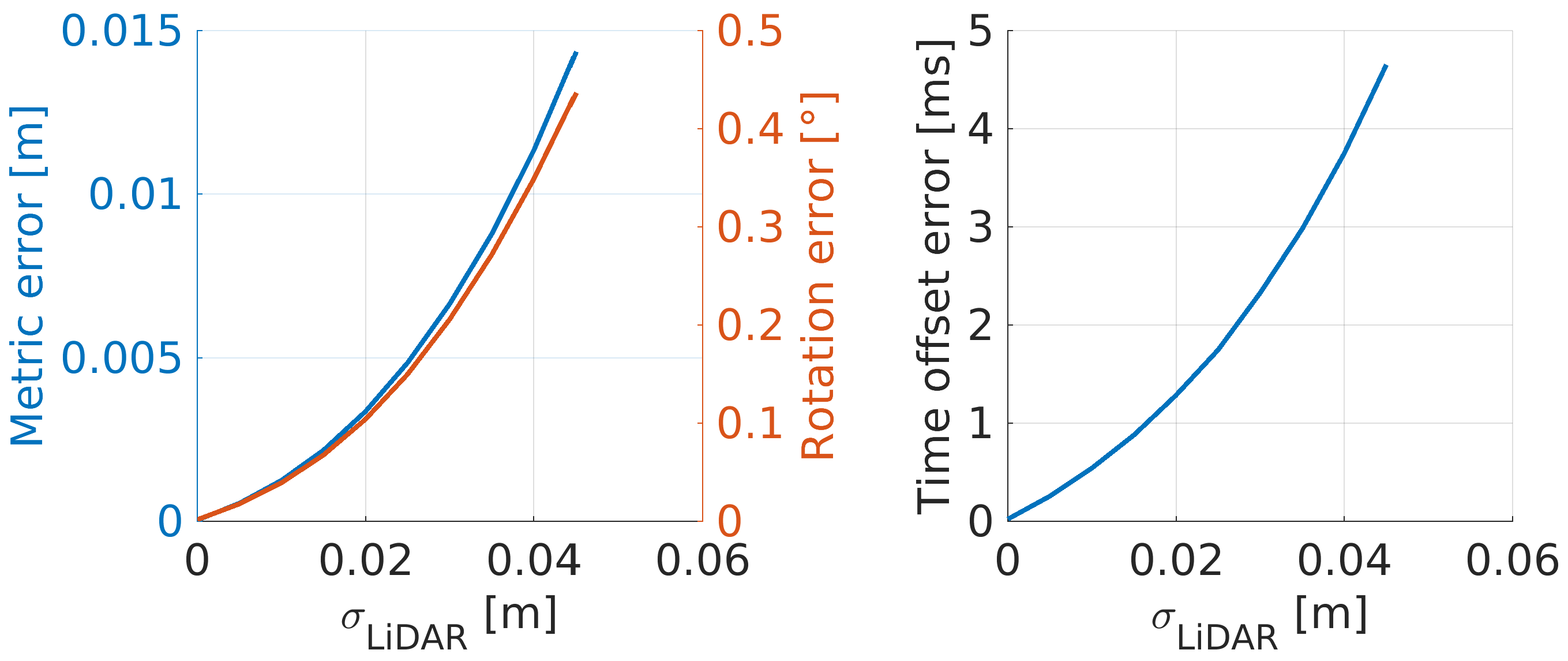}
    \caption{Metric, rotational and time offset errors of the calibration depending on the standard deviation of the LiDAR range measurement ($\sigma_{\rm LiDAR}$)}
    \label{fig:lidarart}
\end{figure}

In the simulation, we assumed that the 3D LiDAR is internally calibrated with a range measurement that has a Gaussian noise~\cite{psicra}.
We performed 1900 random experiments (100 different trajectories for the motion of the calibration marker for
19 different time offsets) for each assumed standard deviation of the Gaussian noise of the LiDAR range measurements ($\sigma_{\rm LiDAR}$).
In each case we measured the metric, rotation, and time offset estimation errors between the real and the estimated values (Fig.~\ref{fig:lidarart}). 

The errors grow significantly when $\sigma_{\rm LiDAR}$ increases with a measured average error of 1.13 cm, 0.35\textdegree~and 3.75 ms when $\sigma_{\rm LiDAR}$ is equal to 0.04 m.
Fortunately, the Velodyne VLP-16 has an accuracy of $\pm 2.5$ cm, which corresponds to the standard deviation of about $0.01$ m.
In such a case, we obtained average calibration errors of 0.12 cm, 0.04\textdegree~and 0.54 ms.

\subsection{Influence of the initial time offset guess}

\begin{figure}[htbp!]
    \centering
    \includegraphics[width=\columnwidth]{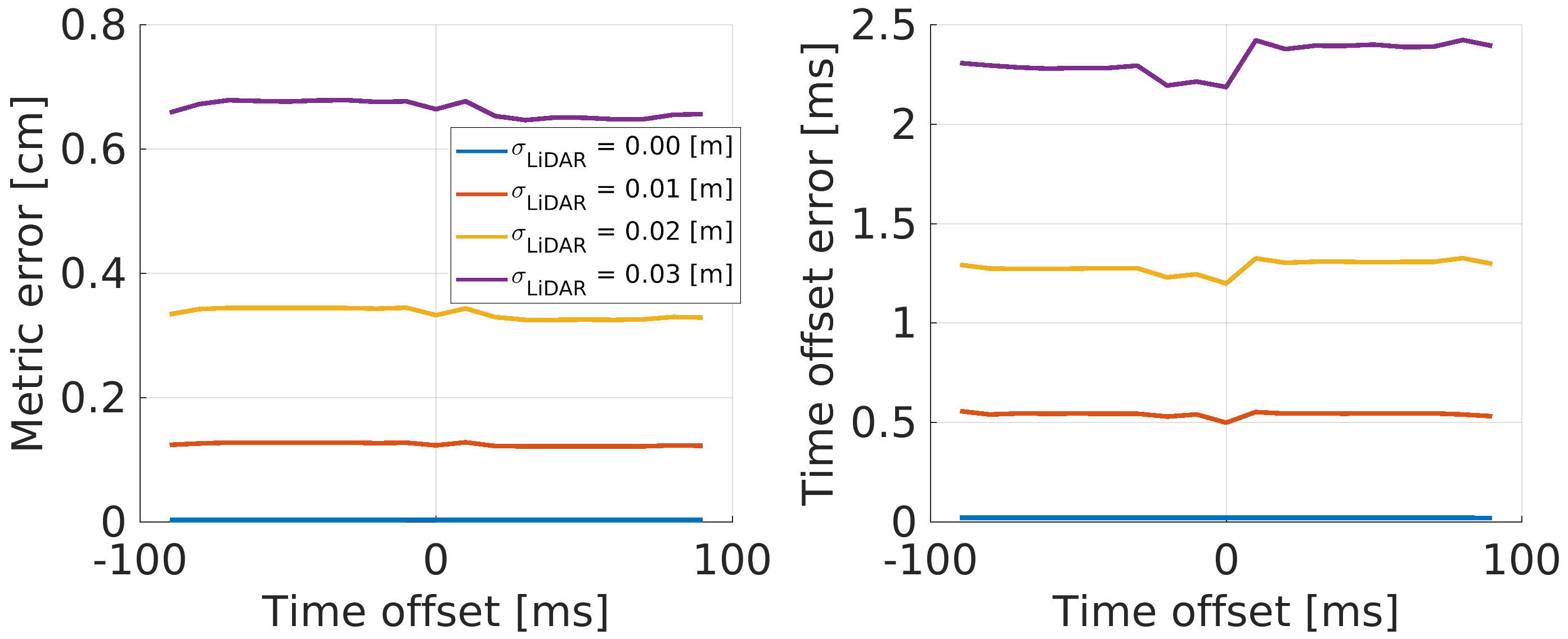}
    \caption{Metric error and time offset error depending on the initial time offset guess and chosen standard deviation of the LiDAR range measurement ($\sigma_{\rm LiDAR}$)}
    \label{fig:inittime}
\end{figure}

A good initial guess is crucial for gradient-based optimization to converge to the global minimum. 
In real-world scenarios, the sensible initial guess for spatial transformation can usually be provided (i.e. by measuring with a tape), which might not be true for temporal calibration (although cross-correlation can be used in many scenarios).
The results for experiments with initial time offsets between sensors that differed from $-90$ ms to $90$ ms from ground truth time offsets are presented in Fig.~\ref{fig:inittime}.

In each considered case, our system converged to similar metric and time offsets regardless of the initial time offset guess and initial guess of the transformation proving empirically that the optimized function is convex.
The metric and time offset errors increase with standard deviation values $\sigma_{\rm LiDAR}$.

\subsection{Spatial vs. spatiotemporal calibration}

\begin{figure}[htbp!]
    \centering
    \includegraphics[width=\columnwidth]{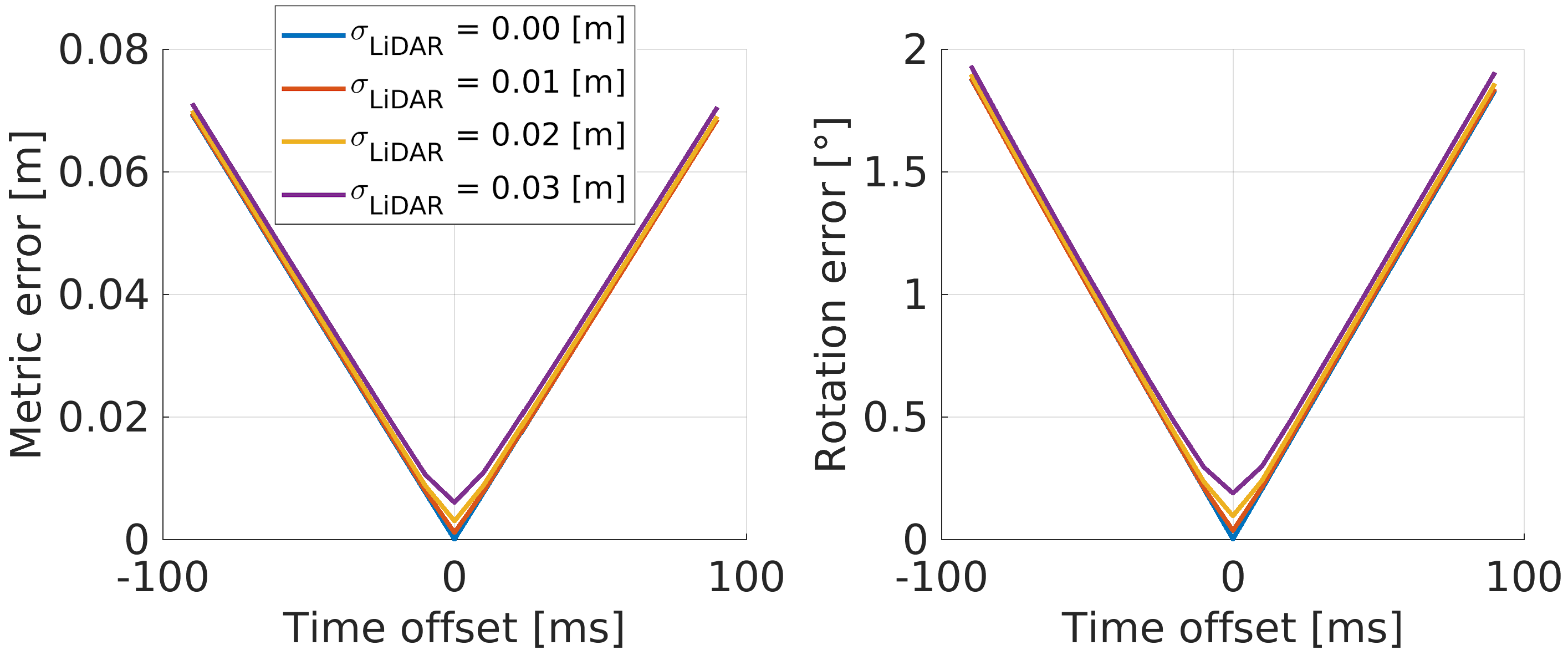}
    \caption{Metric, rotational and time offset errors of the calibration depending on the chosen time offset when only spatial calibration is performed (no temporal calibration)}
    \label{fig:lidartime}
\end{figure}

We performed experiments when only the spatial transformation between sensors was optimized to check what errors are to be expected if no temporal calibration would be performed (Fig.~\ref{fig:lidartime}).

The experiments confirmed that the time offset estimation is crucial as the average metric error of transformation of 3.07 cm and 0.83\textdegree~was measured when the time offset was equal to 40 ms and only increased when the time offset was larger.
Similar errors were observed for all analyzed values of standard deviation $\sigma_{\rm LiDAR}$.
This proves that the users should choose spatiotemporal calibration over spatial calibration in almost all scenarios (with an exception when hardware synchronization is available).

\subsection{Influence of the number of constraints}

\begin{figure}[htbp!]
    \centering
    \includegraphics[width=\columnwidth]{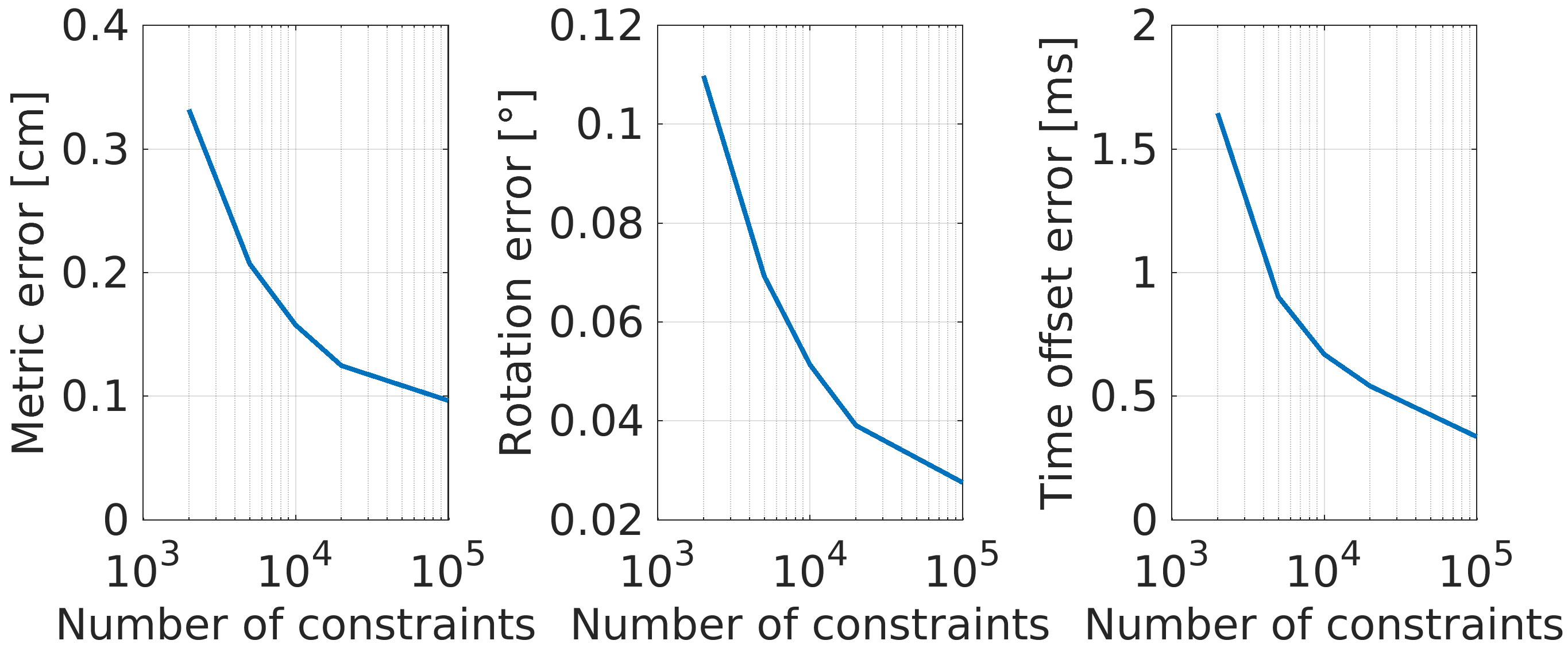}
    \caption{Metric, rotational and time offset errors of the calibration depending on the number of constraints included in the optimization with $\sigma_{\rm LiDAR} = 0.01$ m}
    \label{fig:edge_selection}
\end{figure}

In this experiment, the idea was to verify the accuracy of the obtained results based on the number of LiDAR's point-to-plane constraints included in the optimization (Fig.~\ref{fig:edge_selection}). 

Results indicate that including more constraints (that are different than those already added) reduces the overall metric, rotation, and time offset errors. 
Larger number of constraints increases also the optimization time but still makes it possible to achieve solution in several minutes in worst case scenarios.
What is important, the differences in accuracy between a version with almost all possible constraints (100000) and taking into consideration ten times fewer constraints (10000) is relatively small. 
From these performed experiments we believe that calibration of around 50 seconds, with diverse calibration marker poses in all feasible directions, and rather dynamic motion should be sufficient to obtain accurate calibration under real-world conditions.

\subsection{Influence of the camera framerate}

\begin{figure}[htbp!]
    \centering
    \includegraphics[width=\columnwidth]{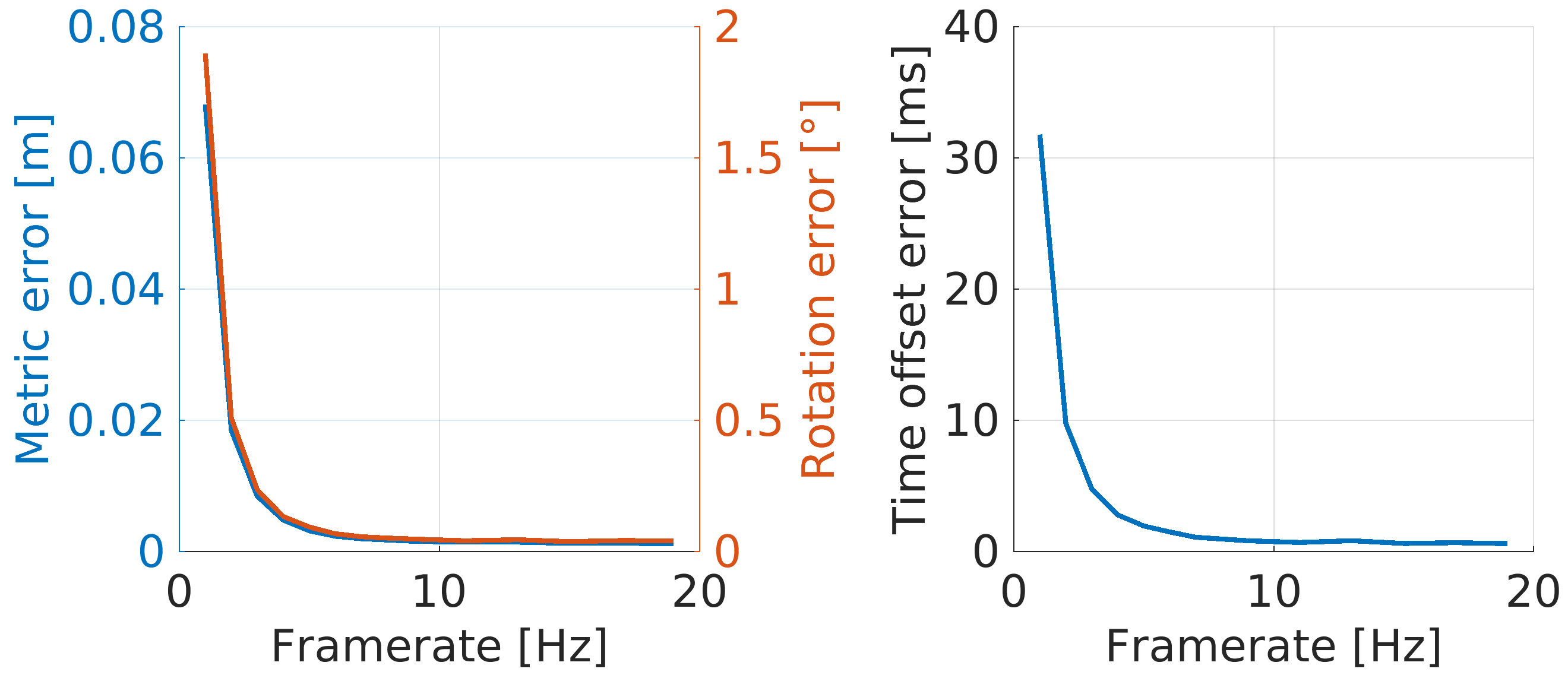}
    \caption{Metric, rotational and time offset errors of the calibration depending on the framerate of camera chessboard detections with $\sigma_{\rm LiDAR} = 0.01$ m}
    \label{fig:framerate}
\end{figure}

The amount of data used in the optimization depends on the framerate of the camera and 3D LiDAR. 
In the case of LiDARs, most of the available sensors provide data with 10 Hz but cameras provide possibilities to acquire images even up to hundreds fps.
Therefore, we verified the accuracy depending on the chosen framerate from the camera (Fig.~\ref{fig:framerate}).

Results showed that there is no visible benefit when the camera framerate exceeds 10 Hz. 
A lower framerate might negatively influence the accuracy of metric, rotation, and time offset estimations, but higher framerates do not improve the accuracy of the calibration.
This proves that the chosen B-spline interpolation works well as long as the interpolation interval does not exceed 100 ms.

\section{Experiments}

\subsection{Velodyne VLP-16 and stereo-camera setup}

\begin{figure}[htbp!]
    \centering
    \includegraphics[width=\columnwidth]{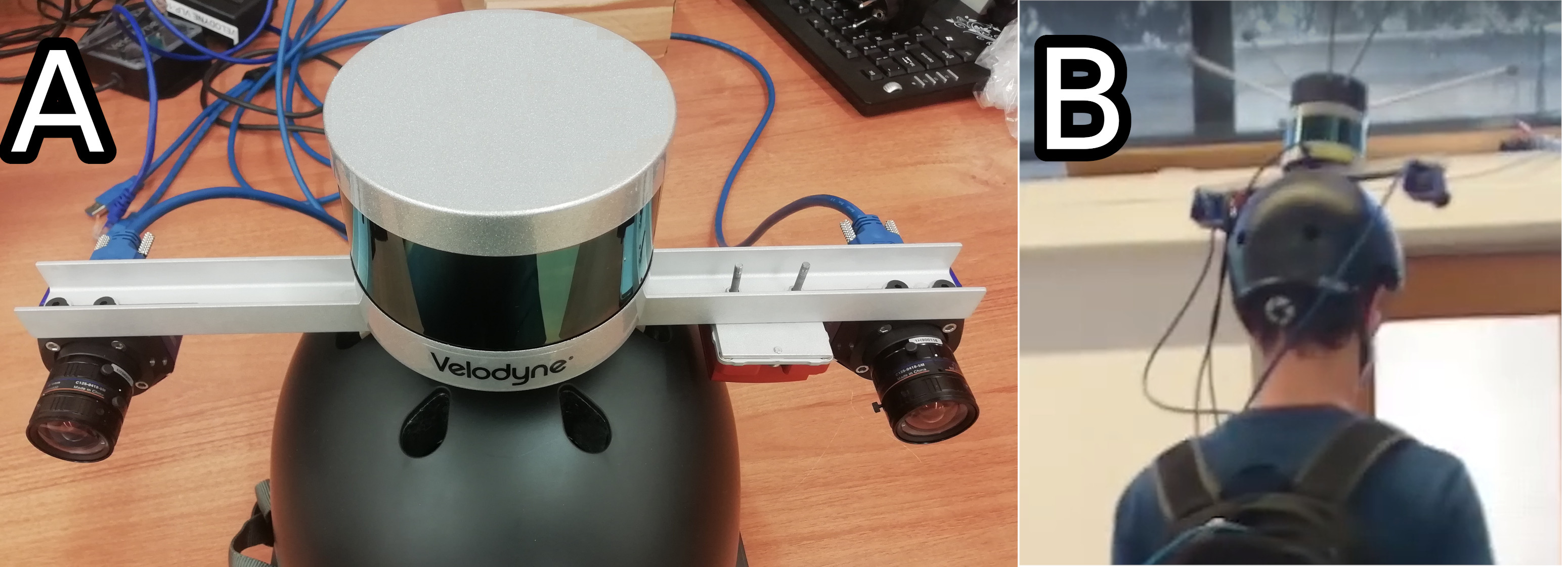}
    \caption{The experimental setup with two MV BlueFox3-2016C-1112 cameras and Velodyne VLP-16 LiDAR without hardware trigger (A) that was used for indoor mapping when mounted on a helmet (B)}
    \label{fig:expvelo}
\end{figure}

The experimental setup for real-world verification comprised of two MV BlueFox3-2016C-1112 global-shutter cameras with Basler Lens C125-0418-5M F1.8 f4mm rigidly mounted on both sides of the Velodyne VLP-16 as presented in Fig.~\ref{fig:expvelo}.
The presented setup lacks hardware trigger, so, the cameras were software triggered with our custom driver at the same time ensuring close (but not perfect) trigger times. 

At first, we calibrated both cameras using \textit{kalibr}~\cite{kalibr1, kalibr2} achieving a transformation from the left camera to the right camera, ${}^{\rm right}\mathbf{K}_{\rm left}$, that was almost identical to manually measured offset with \textit{kalibr} reporting accuracy of $\pm 0.06$ cm and $\pm 0.0991\degree$ and thus we treat it as ground-truth reference. 
Then, we used our software twice to calibrate Velodyne VLP-16 with the left camera (${}^{\rm left} \mathbf{T}_{\rm velo}$), and to calibrate Velodyne VLP-16 with the right camera (${}^{\rm right} \mathbf{T}_{\rm velo}$). 
We compute the difference between both calibrations as:
\begin{equation}
    \Delta \mathbf{T} = {}^{\rm right}\mathbf{K}_{\rm left} {}^{\rm left} \mathbf{T}_{\rm velo} {}^{\rm right} \mathbf{T}_{\rm velo}^{-1}.
\end{equation}
With the original baseline of 31 cm, we measured the translational difference of \textbf{T} to be 0.74 cm and the rotational difference to be equal to 0.97\textdegree. 
We obtained time offsets of $\Delta t_1 = 3.99$ ms and $\Delta t_2 = 3.77$ ms for both camera-LiDAR calibrations.
The difference between $\Delta t_1$ and $\Delta t_2$ can be attributed to imperfect software trigger of our sensory setup solution.

\subsection{MRS6124 and stereo-camera setup}

\begin{figure}[htbp!]
    \centering
    \includegraphics[width=\columnwidth]{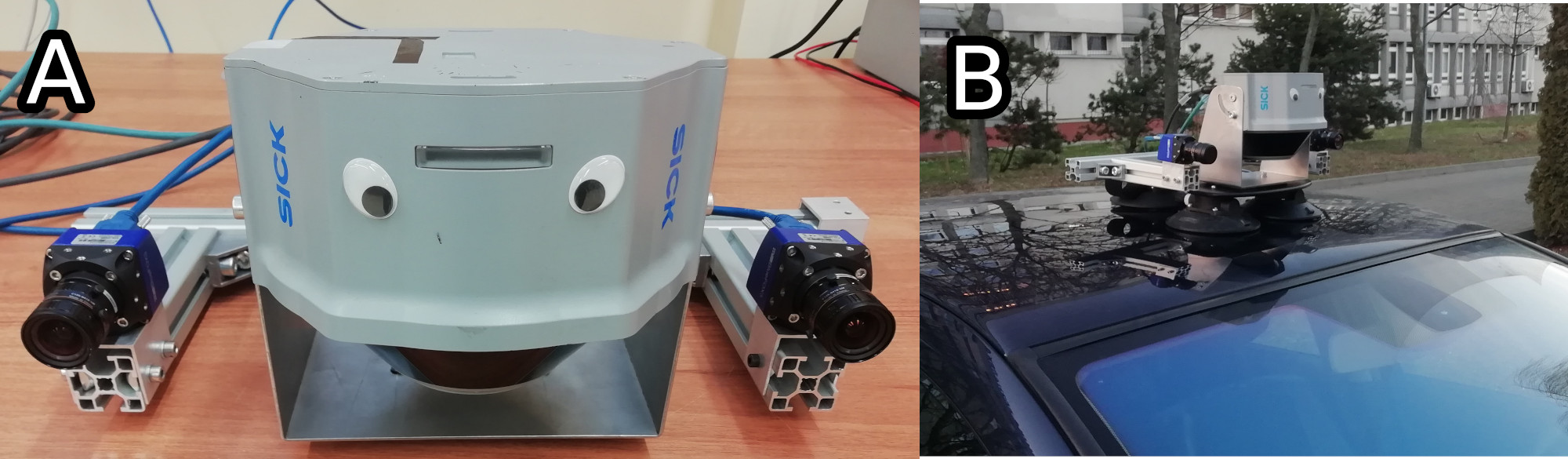}
    \caption{The experimental setup with two MV BlueFox3-2016C-1112 cameras and SICK MRS6124 LiDAR without hardware trigger (A) that is used in automotive applications (B)}
    \label{fig:expmrs}
\end{figure}

A similar experiment was repeated with a different setup consisting of the same cameras and SICK MRS6124 (Fig. \ref{fig:expmrs}), which is a more challenging scenario due to the lower accuracy of SICK MRS6124 (systematic error of $\pm 12.5$ cm and statistical error of $3$ cm) when compared to the Velodyne VLP-16. 
Similarly, we calibrated both cameras using \textit{kalibr} that reported an accuracy of $\pm 0.0666$ cm and $\pm 0.1110\textdegree$, and then verified achieved results with two independent camera-LiDAR calibrations using the proposed software. 
We measured the difference between \textit{kalibr} and our software to be equal to 0.73 cm and 1.45\textdegree~for a baseline of approximately 34 cm. 

\begin{table*}[ht!]
\centering
\caption{Comparison of the reported errors ($t_e$ is translation error, $r_e$ is rotation error) of the proposed solution with respect to the state-of-the-art. Please note that the reported accuracy for each method were NOT obtained in the same conditions and vary depending on the source. The parameter type informs about the calibration: S is spatial, S \& T is disjointed spatial and temporal calibration while ST stands for spatiotemporal calibration. HDL-64E stands for Velodyne HDL-64E, VLP-16 for Velodyne VLP-16.}
\label{tab:acc}
\begin{tabular}{ccccccc} \hline
  Algorithm &  CalibNet~[27] & Taylor \textit{et al}.~[10] & Park \textit{et al}.~[13] & Taylor \textit{et al}.~[10] & Ahmad \textit{et al}.~[26] &  Ours \\   \hline
  Source & [27] & [10] & [13] & reported in [13] & reported in [13] & experiment \\ 
  Type of & S & S \& T & ST & S \& T & S & ST \\ 
  LiDAR & HDL-64E & HDL-64E & VLP-16 & VLP-16 & VLP-16 & VLP-16 \\  \hline
        $t_e$ [m]   & 0.149 & 0.099 & 0.01  & 0.15          & 0.12          & \textbf{0.0043} \\
$r_e$ [\textdegree] & 0.93 & 0.622  & 0.4 &   2.6            & 4.59            & \textbf{0.33} \\ \hline

\end{tabular}
\end{table*}

\begin{figure}[htbp!]
    \centering
    \includegraphics[width=\columnwidth]{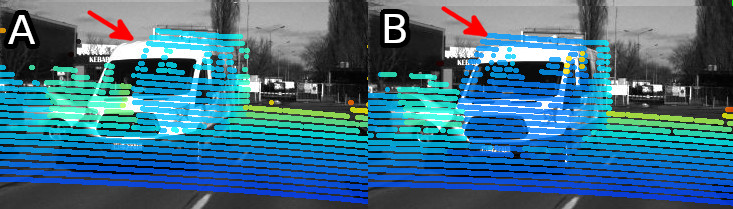}
     \caption{Visual verification of the calibration of the sensory setup with MRS6124 in the cases of spatial (A) and spatiotemporal (B) calibration. Notice the difference for a moving car marked by red arrows}
    \label{fig:mrsverif}
\end{figure}

Our calibration indicated that the measured timestamps have to be corrected by $200.99$ ms and $202.41$ ms for the left and right camera, respectively.
In our opinion, such a significant time offset stems from the buffering of data either in the sensor itself or in the ROS driver.
The visualization confirms that LiDAR points are significantly delayed when compared to camera images (as presented in Fig.~\ref{fig:mrsverif}). 
When no temporal calibration would be performed (time offset parameter fixed to 0 in the optimization), we achieved a difference to \textit{kalibr} of 7.51 cm and 4.97\textdegree.
This experiment confirms the usefulness of our contribution as the time offset was coherently estimated for both camera-LiDAR calibrations and seems correct when visually inspected.

\subsection{Comparison with state-of-the-art}

Fair comparison of spatial or spatiotemporal calibration solutions is challenging as each method has its own, different requirements, and the best performance can be achieved in certain conditions.
The accuracy of the calibration strongly depends on the accuracy of the used sensors and the amount of data it provides (in practice the number of scan lines and the angular field of view).
Moreover, obtaining precise enough ground-truth for the 7 DoF spatial and temporal relations in order to asses the calibration accuracy is challenging.
Nevertheless, we decided to perform extensive evaluation of the accuracy of the proposed solution and then provide an overall comparison with the existing state-of-the-art solutions.


The accuracy of our solution was measured for the stereocamera setups with Velodyne VLP-16 and SICK MRS6124. 
Similarly to results reported in previous section we assumed left-right camera calibration from \textit{kalibr} as the ground-truth and compared our results to that result. 
We run our calibration software 50 times on real data by randomly selecting $35\%$ of all available constraints for both left camera - LiDAR and right camera - LiDAR calibrations achieving 2500 different calibrations to compare to results reported by~\textit{kalibr}. 
The obtained results are presented in Table~\ref{tab:acc}.

The results obtained by our system for Velodyne sensor show that the proposed system is accurate.
The obtained accuracy is comparable to the best reported results by Park~[13] but as both reported results used different sources of ground truth (Park used manual ground truth) we would abstain from claiming the best accuracy.
From our experiments we also computed the standard deviation of the calibration, that was equal to $\sigma_t = 0.03$ cm and $\sigma_r = 0.005\degree$~for translational and rotational parts, respectively.

When it comes to SICK MRS6124, we performed the same verification procedure, however, the results cannot be directly compared to the
results from~\cite{park20} and other works using Velodyne LiDARs, due to
significantly different measurement characteristics of the SICK LiDAR.
Nevertheless, the MRS6124 results demonstrate that the proposed calibration framework is feasible for very different types of LiDARs.
Calibrating the SICK sensor we obtained similar translational error $t_e = 0.8$ cm with significantly greater rotational error $t_r = 3.52\degree$.
This is the result of two factors: firstly, MRS6124 has a significantly worse accuracy than Velodyne, and secondly, the laser beams are more tightly packed.  
Both of this factors combined make it difficult to properly constrain the rotational part of the calibration.
We believe that using a larger calibration pattern and recording calibration data with more extreme orientations of the calibration pattern should be sufficient to overcome this issue.

\section{Conclusions}

We propose a novel camera-LiDAR calibration software that is the first marker-based solution that provides spatiotemporal calibration owing to the novel B-spline interpolation of plane equations employing a minimal plane representation in Lie algebra.
The solution requires only a commonly available calibration marker and a short, one-minute calibration session to provide repeatable and accurate results. 
Moreover, it is available as a ROS-compatible package.

The simulation experiments over a significant number of random scenarios show that we should expect a centimeter-level accuracy when using currently available 3D LiDARs, while demonstrating also, that our solution is robust to bad spatial and temporal initial guesses.
We performed two real-world experiments with Velodyne VLP-16 and SICK MRS6124 LiDARs. The indirect comparison with \textit{kalibr} shows that our method provides results of similar accuracy. 
This suggests that the presented calibration procedure can be practically combined with \textit{kalibr} in the ROS environment whenever calibration of a more complicated sensory system with a 3D LiDAR and a camera is necessary.
The experimental results demonstrate also that the temporal calibration is critical for applications and should be performed when no hardware synchronization is available.




\end{document}